\documentclass{bmvc2k}
\usepackage{amsfonts}
\usepackage{float}
\usepackage{graphicx}
\usepackage{xcolor, soul}
\sethlcolor{green}

\title{G-CMP: Graph-enhanced Contextual Matrix Profile for unsupervised anomaly detection in sensor-based remote health monitoring}

\addauthor{Nivedita Bijlani}{n.bijlani@surrey.ac.uk}{1}
\addauthor{Oscar Mendez Maldonado}{o.mendez@surrey.ac.uk}{1}
\addauthor{Samaneh Kouchaki}{samaneh.kouchaki@surrey.ac.uk}{1}

\addinstitution{
 Centre for Vision, Speech and Signal Processing (CVSSP),\\
 University of Surrey,\\
 Guildford, UK
}

\runninghead{BIJLANI ET AL.}{G-CMP: GRAPH-ENHANCED CONTEXTUAL MATRIX PROFILE}

\def\etal{\emph{et al}\bmvaOneDot}

\begin{document}
\maketitle

\begin{abstract} Sensor-based remote health monitoring is used in industrial, urban and healthcare settings to monitor ongoing operation of equipment and human health. An important aim is to intervene early if anomalous events or adverse health is detected. In the wild, these anomaly detection approaches are challenged by noise, label scarcity, high dimensionality, explainability and wide variability in operating environments. The Contextual Matrix Profile (CMP) is a configurable 2-dimensional version of the Matrix Profile (MP) that uses the distance matrix of all subsequences of a time series to discover patterns and anomalies. The CMP is shown to enhance the effectiveness of the MP and other SOTA methods at detecting, visualising and interpreting true anomalies in noisy real world data from different domains. It excels at zooming out and identifying temporal patterns at configurable time scales. However, the CMP does not address cross-sensor information, and cannot scale to high dimensional data. We propose a novel, self-supervised graph-based approach for temporal anomaly detection that works on context graphs generated from the CMP distance matrix. The learned graph embeddings encode the anomalous nature of a time context. In addition, we evaluate other graph outlier algorithms for the same task. Given our pipeline is modular, graph construction, generation of graph embeddings, and pattern recognition logic can all be chosen based on the specific pattern detection application.We verified the effectiveness of graph-based anomaly detection and compared it with the CMP and 3 state-of-the art methods on two real-world healthcare datasets with different anomalies. Our proposed method demonstrated better recall, alert rate and generalisability. \end{abstract}

\section{Introduction}

\label{sec:intro}
Sensor-based remote health monitoring continues to grow rapidly in a variety of industrial, urban and healthcare settings. Drawing insights from sensor-based data enables the analysis of temporal patterns and detection of adverse conditions, with minimal intrusion and low cost. For example, sensor data related to movement, physiology, behaviour and sleep can be used to gain insights into elderly patients’ health and monitor their condition, allowing for timely assistance and enabling them to live independently for longer. However, real-world sensor-based health monitoring poses unique challenges. It is characterised by multivariate data, absence of reliable labelling as annotation is resource-intensive, data drift, noise and lack of periodicity. An anomaly detection algorithm in the wild must address these issues, and be able to make predictions with low latency, work with personalised baselines (especially in healthcare), deal with irregularly reported data, demonstrate high sensitivity at a low alert rate, require minimal tuning, and be explainable to monitoring teams.
\begin{figure}
\centering
\includegraphics[width=0.95\linewidth]{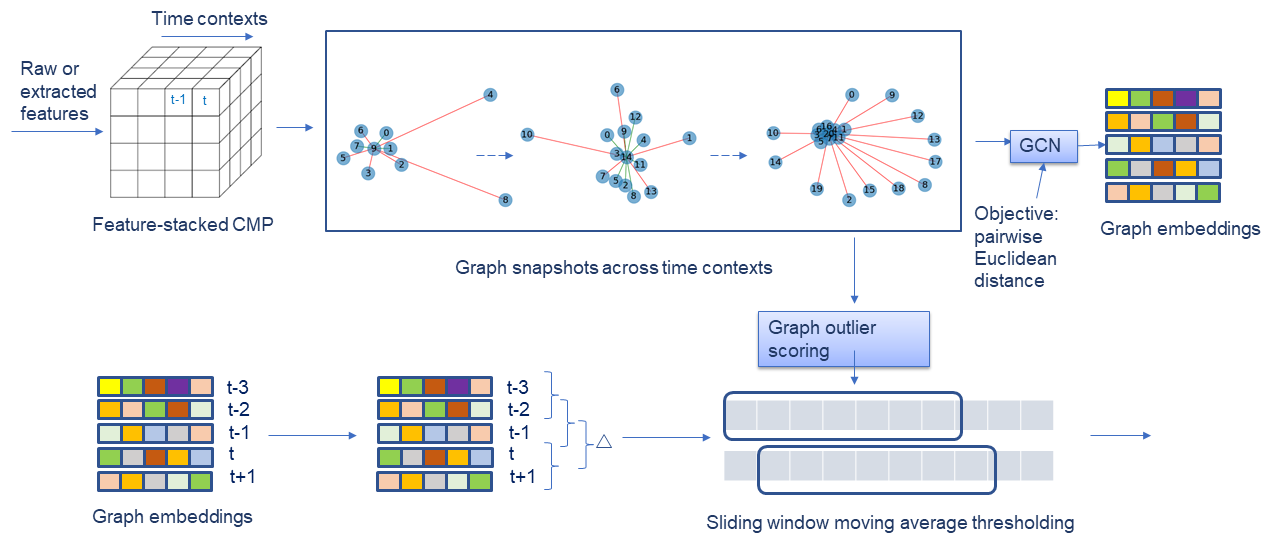}
\caption{Input features are used to create the CMP. The CMP distance matrix is used to create temporal graph snapshots. The central node of each graph represents the time context of interest. A spatial GCN aggregates feature information from previous nodes to produce graph-level embeddings. Successive graph embeddings are compared, and sliding window moving average thresholding is used to detect anomalous time contexts or adverse events.}\vspace{-0.5cm}
\label{fig:pipeline}
\end{figure}
In this paper, we focus on fast, lightweight self-supervised anomaly detection that is robust to the noisy data and labels common in sensor-based remote health monitoring. Our smart healthcare approach uses personalised baseline data for adaptive anomaly detection. We propose and evaluate a solution based on the Matrix Profile (MP) \cite{yeh2016matrix} - a modern, exact, and ultra-fast distance-based anomaly detection algorithm - specifically its recent, more flexible variant, the Contextual Matrix Profile (CMP) \cite{DEPAEPE2020103487}. The idea of CMP is rooted in the intuitive visualization of patterns and anomalies. By aggregating noisy multivariate sensor observations into time blocks or ``contexts'' the CMP makes it easier to distinguish normal from anomalous data. Existing work based on the CMP is currently limited in its ability to scale to high dimensions, or address cross-sensor correlations \cite{bijlani2022unsupervised}. Our approach overcomes these limitations using graph-based machine learning. Specifically, we use the CMP distance matrices to construct ``time context graphs'', and evaluate each graph in relation to previous graphs by applying self-supervised graph models. We then use the individual graph embeddings to uncover spatiotemporal anomalies using a sliding window approach. The speed of our approach stems from two areas: (1) the use of MP which internally uses the Fast Fourier Transform for distance computation, and (2) the use of one-hop graphs that harness the power of graph representation learning but with small parameter size and computational complexity. Our works build upon the work of~\cite{bijlani2022unsupervised} which demonstrates the effectiveness of the CMP-based approach in unsupervised anomaly detection. Using graph models on top of the CMP additionally opens up the opportunity for the generated embeddings to be used by downstream algorithms to understand and distinguish between specific anomaly conditions. Our approach is realised via a modular pipeline as shown in Figure \ref{fig:pipeline}. We apply our algorithm to detect anomalous activity indicative of adverse health conditions in 2 real-world healthcare datasets averaging over 7000 days of elderly patient activity data, and obtain better sensitivity, and lower alert rate relative to the multivariate CMP as well as 3 benchmark anomaly detection methods. Our work makes the following contributions:\vspace{-0.125cm}
\begin{itemize}
  \item Combines graph models with CMPs for adaptive anomaly detection.\vspace{-0.125cm}
  \item Enhances the potential and efficacy over current CMP-based anomaly scoring in a remote healthcare monitoring scenario.\vspace{-0.125cm}
   \item Proposes a flexible and novel end-to-end unsupervised anomaly detection pipeline for remote health monitoring in noisy settings (see Figure \ref{fig:pipeline}).
\end{itemize}

\section{Related Work}
\label{sec:relatedwork}
In a dynamic graph environment, the aim is to identify abnormal graph snapshots based on unusual evolution of features ~\cite{Ma_2021}. In order to create suitable embeddings, a guiding function or self-defined labels are required. Teng \etal \cite{teng2018deep} combine LSTM autoencoders with hypersphere learning to capture graph temporal features and separate normal from abnormal graph snapshots based on the learned hypersphere and magnitude of reconstruction error. However, the degree of anomaly pollution and noise can degrade its performance, and a continual learning strategy is not addressed. Yang \etal~\cite{yang2022flow} use a negative sampling strategy to create a variational autoencoder which is used to guide the training of an autoencoder optimizer, enabling it to distinguish between normal and abnormal samples. This method utilizes node connectivity only, ignoring node features. Mo \etal ~\cite{mo2022simple} use contrastive learning to minimise intra-class variation and maximise inter-class variation. While this removes the need for any discriminator or data augmentation and improves speed and scalability, it is unclear how anchor embeddings can be updated in a temporal context, and if contrastive loss can be reliably applied to noisy data. Zheng \etal ~\cite{zheng2019addgraph} introduce the AddGraph framework which uses a temporal GCN with an attention-based GRU to combine the hidden states for long-term behaviour patterns and window information containing the short-term patterns of the nodes. The hidden state of the nodes at each timestamp are used calculate the anomalous probabilities of existing edges and negative sampled edges in future snapshots, and fed to a margin loss. While this approach addresses noise and label issues with data, it assumes that training data set is largely normal, and does not address training data refresh. Cui \etal ~\cite{cui2020adaptive} apply an adaptive learning strategy via a Laplacian smoothing filter to smooth node features and constructs positive training examples as node pairs that are adjacent and also have similar smoothed node features. Node pairs with lowest similarity are selected as negative pairs and the model is trained via cross-entropy loss. This method has high computational complexity and low scalability due to pairwise similarity computations, limiting its usefulness in a streaming data scenario.
\section{Approach}
\label{sec:approach}

\begin{figure}
\centering
\includegraphics[width=0.7\linewidth,keepaspectratio]{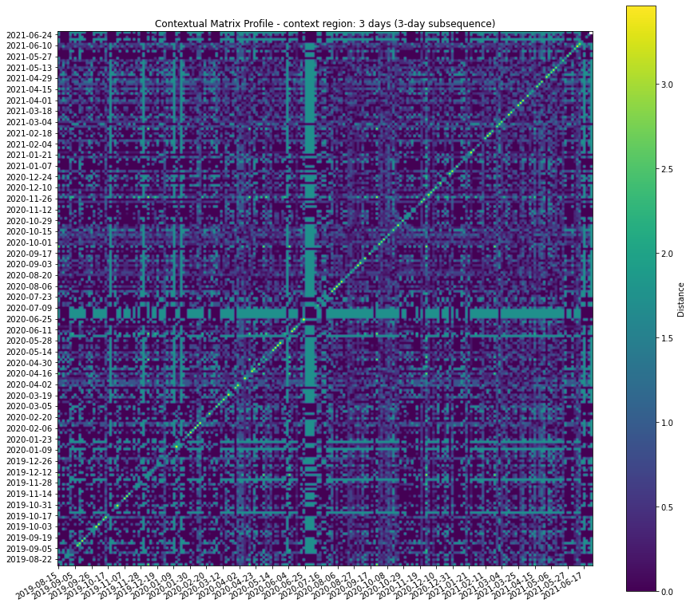}
\caption{CMP of daily late evening bathroom activity for a participant. Green bands indicate anomalous activity. The CMP is symmetric around the diagonal.}\vspace{-0.25cm}
\label{fig:cmpthumbnail}
\end{figure}

\begin{figure}
\centering
\includegraphics[width=0.9\linewidth,keepaspectratio]{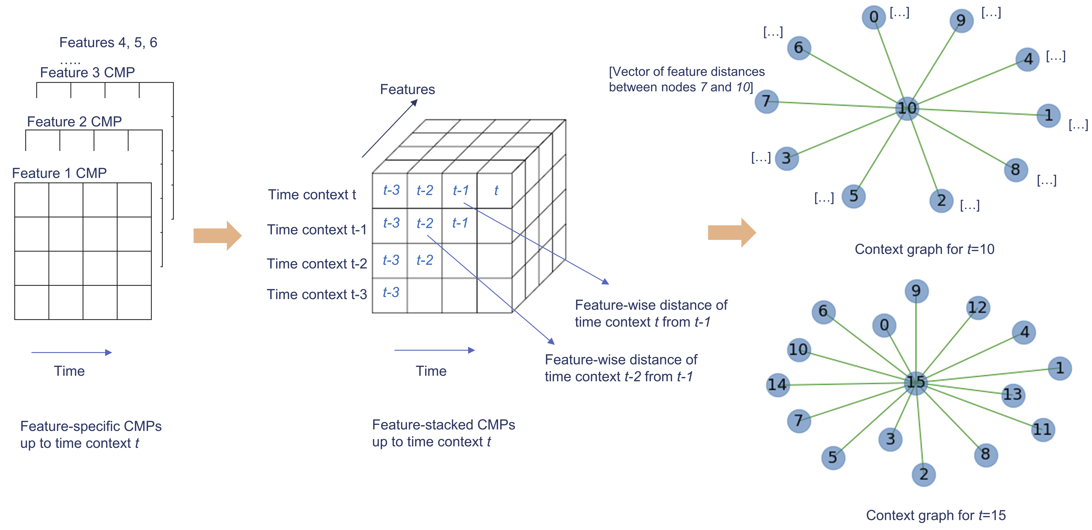}
\caption{Each cell in the Contextual Matrix Profile is converted to a star graph}\vspace{-0.25cm}
\label{fig:cmpgraph}
\end{figure}

The Contextual Matrix Profile (CMP) \cite{DEPAEPE2020103487} is a distance-based approach built on the Matrix Profile (MP), a state-of-the-art time series analysis technique used for pattern detection, anomaly detection, time series segmentation and change point detection \cite{yeh2016matrix}. The CMP is a configurable, 2-dimensional version of the MP that tracks the minimum distance between each context or block of subsequences in user-defined regions of the time series, which forms one cell in the CMP (Figure \ref{fig:cmpthumbnail}). The CMP acts to denoise the time series, allows focus on specific regions of interest, and fixes the anomaly masking problem seen in the MP. Recently, \cite{bijlani2022unsupervised} developed the multidimensional CMP for multivariate time series data and offered a real-world use case in healthcare anomaly detection. By averaging the nearest neighbour distance to previous time contexts, the authors generated context-wise anomaly scores and used statistical thresholding to detect anomalies.
Our approach seeks to combine the power of unsupervised graph representation learning with the speed, simplicity and inherent interpretability of the CMP. We consider each time context or cell of the CMP as a star graph where the central node represents the time context under consideration, and outer nodes represent previous time contexts. The intuition is that the anomalous nature of a context is relative to the state of previous contexts. We describe each outer node by a vector of feature distances from the central node taken from the feature CMPs as illustrated in Figure \ref{fig:cmpgraph}. Thus, we model successive time contexts as a growing temporal set of star graphs. We then apply unsupervised GNN and other graph outlier techniques to detect anomalies, as shown in Figure \ref{fig:pipeline}. The addition of graph-based anomaly detection has important advantages - it enables automatic feature extraction from raw features, scales easily when new features are added -  the structure of the graph does not change, only the node feature vector expands - and enables easy visualisation and interpretability.

\subsection{Graph-based anomaly detection}
Figure \ref{fig:pipeline} shows two different ways of detecting anomalies from the temporal graph snapshots. The first approach uses a self-supervised graph convolutional network to generate embeddings for each context graph. We compute the distance between consecutive graph embeddings, and these deltas are subject to 7-context sliding window-based thresholding. A time context is identified as an anomaly if its delta value is greater than the threshold of 1 standard deviation around the sliding window moving average. The second approach applies graph outlier scoring to the temporal stream of graphs using four different algorithms and applies similar sliding window based thresholding to successive graph outlier scores. The biggest benefit of using graph-based anomaly detection as described is modularity. 

Unlike standard CMP-based approaches, our novel GNN pipeline offers flexibility common in Machine Learning approaches, such as adjusting the number and type of input features, which can be engineered or automatically extracted. The temporal graphs can also be constructed in different ways. Similarly, the self-supervised objective can be easily replaced, and the choice of the embedding distance measure can be changed, or an attention network might be slotted in to weight embeddings or scores. Likewise, the thresholding criteria can be configured according to the domain. Furthermore, in this work, we demonstrate the strength of combined approaches by combining CMP-based distance metrics with different GNN architectures, as shown in section \ref{sec:gos}. Fundamentally, this demonstrates that CMPs may offer a viable foundation for graph-based anomaly detection.

\subsubsection{Self-supervised single-layer GCN}

\begin{figure}
\centering
\includegraphics[width=0.8\linewidth,keepaspectratio]{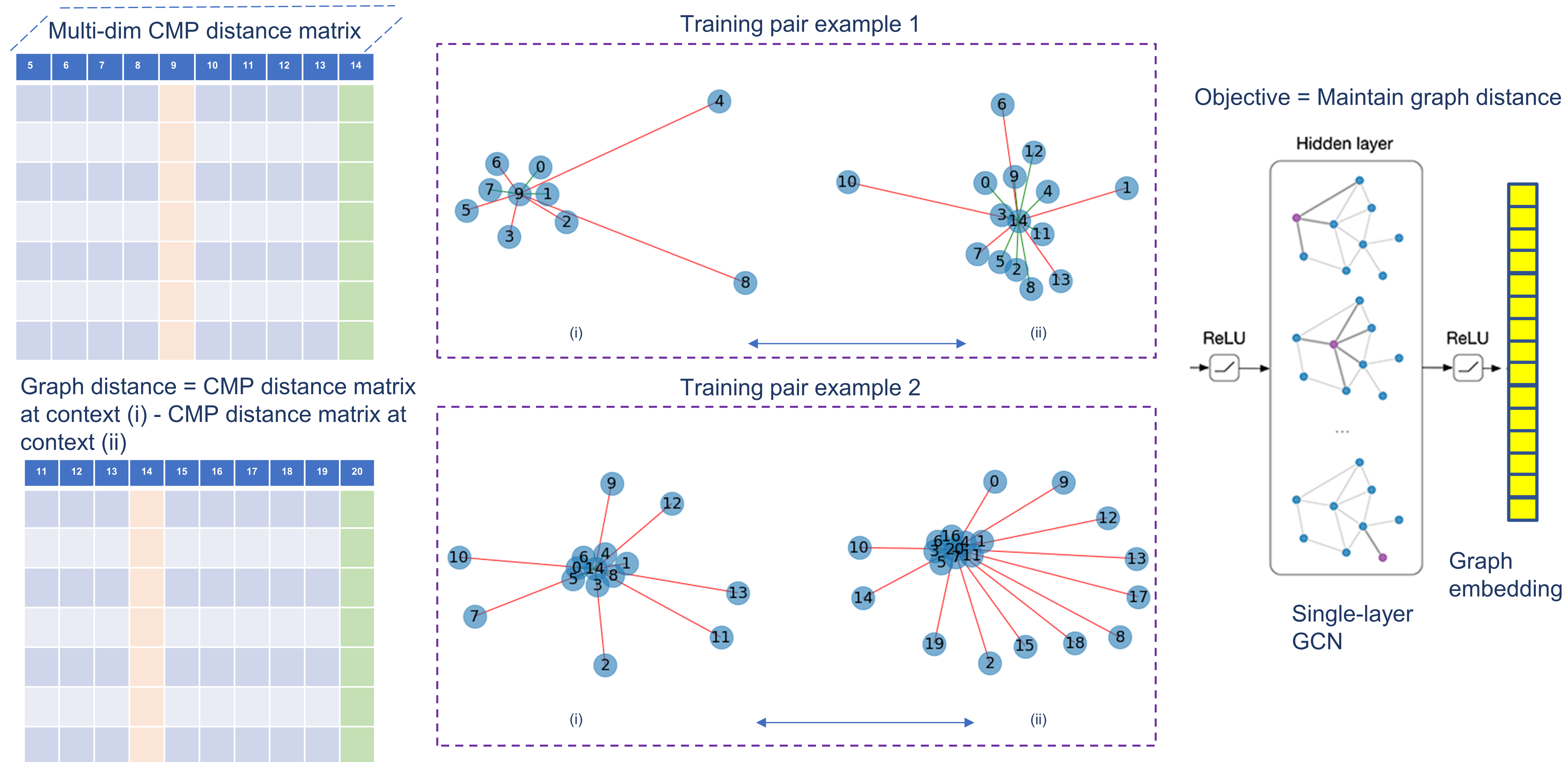}
\caption{A single-layer GCN is trained on distance between graph pairs. Distance is measured as the difference in entropy between the corresponding time contexts as calculated from the CMP distance matrix.}\vspace{-0.25cm}
\label{fig:gcnemb}
\end{figure}

A graph convolutional neural network is a model that learns a function over the structural features and attributes of a graph, to produce a node-level or graph-level output representation. The same convolutional weights or filters are shared by the entire graph. The point of graph convolution is to aggregate node and edge information from node neighborhoods to extract local information\cite{kipf2016semi}. To extract multi-scale substructure features, multiple graph convolution layers may be stacked \cite{zhang2018end}.
In our approach to anomaly detection, we utlize a spatial GCN to evaluate the anomalous nature of a time context (the central node in the star graph) in terms of the features of the preceding time context nodes (outer nodes in the star graph) it is connected to. A single layer GCN is sufficient because each context graph is a simple one-hop graph where the central node (time context being evaluated) is only one hop away from its outer nodes. We use unsupervised graph embeddings based on graph distance introduced in \cite{bai2019unsupervised} and apply the intuition that an anomaly represents an increase in the energy of the system. Hence, separating pairs of graphs based on the difference in their energy might be an effective way to separate low-energy (regular) from high-energy (anomalous) nodes. Thus, our GCN is trained using random pairs of graphs as input, with the goal of learning the energy difference function. We calculate this as the Frobenius norm between the CMP distance matrix up to the context represented by the first graph and the context represented by the second graph.
\subsubsection{Graph outlier scoring}
\label{sec:gos}
We adapt four unsupervised graph-based detection algorithms to the problem of graph outlier detection in sensor-based remote health monitoring. Each of these algorithms generates outlier scores, and we apply sliding window based thresholding to discover the anomalous graph in the temporal stream of graphs.

Firstly, we modify Deep Anomaly Detection on Attributed Networks (DOMINANT) \cite{ding2019deep}, a GCN-based autoencoder that models node interactions with layers of nonlinear transformations, and enables the detection of anomalous nodes using a deep autoencoder framework to reconstruct the original attributed network with the learned node embeddings. The reconstruction errors of nodes are employed to flag anomalies. The algorithm allows the user to assign different weights to structure and attribute reconstruction depending on the problem domain. In our case, as the structure of graphs is uniform, we specify \begin{math} \alpha = 0.9 \end{math} to assign 90\% weight to node attribute information. Secondly, the Multilayer Perceptron-based autoencoder (MLPAE) \cite{sakurada2014anomaly} uses a denoising autoencoder to extract latent features from input data via non-linear dimensionality reduction and reconstruct the original data. The reconstruction error gives the anomaly score. We use this method for attribute reconstruction on individual graph nodes. Thirdly, One-class Graph Neural Network (OCGNN) \cite{wang2021one} is a hypersphere learning framework that combines the powerful node representation ability of GNNs with the hypersphere learning objective to detect anomalies. The GNN computes node embeddings by aggregating node neighbourhood information, while the hypersphere learning objective acts to separate normal nodes from anomalous nodes. The anomaly score measures the distance of the embedding with respect to the sphere. Finally, the GCN Autoencoder \cite{kipf2016variational} is a simple but powerful method of learning interpretable latent graph representations, combining a GCN encoder and an inner product decoder applied to the latent variables. The anomaly score reflects the reconstruction error from the variational autoencoder. These four models present a snapshot of SOTA GNN methods for self-supervised learning. Our novel modular approach is capable of integrating seamlessly with these models, demonstrating the flexibility of our approach to self-supervised anomaly detection.\vspace{-0.25cm}

\section {Experiments}
\subsection{Datasets and evaluation criteria}
We evaluate our proposed method on two real-world sensor-based remote health monitoring datasets collected from the homes of persons living with dementia between August 2019 and April 2022, by the UK Dementia Research Institute Care Research and Technology Centre \cite{UKDRI}. We use household movement time series sensor data to detect anomalies related to activity, and correlate these to adverse clinical events, namely agitation and falls. The details of these datasets are summarised in Table \ref{tab:data}.

\begin{table}
\begin{center}
\begin{tabular}{|l|c|c|c|}
\hline
Dataset & Subjects & Days & Anomalies (adverse events) \\
\hline\hline
Agitation cohort & 42 & 13426 & 145\\
Falls cohort & 23 & 5284 & 38 \\
\hline
\end{tabular}
\end{center}
\caption{Details of datasets used}
\label{tab:data}
\end{table}

\begin{table}
\centering
\begin{tabular}{|l|c|}
\hline
Setting & Value \\
\hline\hline
Binning applied to CMP distance matrix & Yes, No\\
Embedding dimension & 64, 128, 256, 512\\
Layer sizes for graph outlier scoring algorithms & 1, 2, 3, 4, 8\\
Dropout & 0.1, 0.2, 0.3\\
Graph distance algorithms &  Entropy, Euclidean\\
Embedding difference algorithms & Chebyshev, Cosine, Euclidean\\
\hline
\end{tabular}\vspace{0.25cm}
\caption{Experimental parameters}
\end{table}

We use three evaluation metrics: sensitivity, average alert rate, and number of subjects for which the model produces a sensitivity greater than $x\%$ ($x$ depends on the noise level of the dataset). The latter metric underscores the generalisability of the model across different households. Due to limited and noisy annotation, precision would be a misleading metric here. We compare the performance of our graph-based models with the univariate and multivariate CMP-based models and three state-of-the-art anomaly detection methods - ABOD (Angle-based outliers)\cite{kriegel2008angle}, COPOD (Copula-based outliers)\cite{li2020copod} and LODA (Lightweight detector of anomalies)\cite{pevny2016loda} included in \cite{bijlani2022unsupervised}. \vspace{-0.125cm}

\subsection{Implementation details}\vspace{-0.125cm}
We aggregate daily the household movement data captured via passive infrared motion sensors, to reduce noise. Successive firings of the same sensor are deduplicated. For each household location, we consider: (1) Total count of sensor firings (2) Early AM (midnight to 6 AM) and late PM (6 PM to midnight) counts (3) Duration at location (4) WS distance between the current and previous day hourly sensor count distributions. A larger WS distance implies a greater change in hourly pattern from one day to the next. This measure is robust to motion densities across households. Several observational studies have shown that erratic bathroom activity, disturbed sleep, agitation and wandering at unusual hours are common characteristics in people with dementia \cite{agata2013challenges} \cite{hu2020unsupervised} \cite{ijaopo2017dementia} \cite{jakkula2007temporal} \cite{rantz2012automated}. All experiments are run on a 64-bit Intel i7-8700K CPU, 3.7 GHz Windows 10 machine with 32 GB RAM. Our self-supervised GCN model is trained to minimise MSE between actual and predicted graph distance, and uses the Adam optimizer with a learning rate of 1e-2, trained for 50 epochs with early stopping (patience=10). The CMP context window size is set to 3, subsequence length to 3 days, sliding window for thresholding to 7 days, and alpha for the DOMINANT algorithm to 0.9. The soft margin for label validation is -10 days, +7 days of the actual label. We tune the hyperparameters (layers, embedding dimension, dropout and choice of distance metrics) based on 5-fold cross-validation using 15 subjects from a separate patient dataset with urinary tract infection (UTI) as the anomalous event. Our experimental settings are shown in Table 2. 

\subsection{Experiment Results}
\begin{table}
\centering
\resizebox{0.8\textwidth}{!}{%
\begin{tabular}{|p{0.6\linewidth} | p{0.08\linewidth} | p{0.07\linewidth} | p{0.09\linewidth} | }
\hline
Model & Recall\% & Alert rate\% & Patient validity\\
\hline\hline
DOMINANT (alpha=0.9, binned) & 73.62 & 5.91 & 36/42\\
\hline
\textbf{GNN (Euclidean-Cosine, dim=128, binned)} & \textbf{73.97} & \textbf{5.66} & \textbf{37/42}\\
\hline
GCNAE (dropout=0.3, dim=128, layers=4) & 67.38 & 5.41 & 31/42\\
\hline
MLPAE (dropout=0.1, dim=128, layers=1, binned) & 60.57 & 5.52 & 28/42\\
\hline
OCGNN (dropout=0.3, dim=128, layers=1, binned) & 63.80 & 5.76 & 29/42\\
\hline
\hline
COPOD(IQR=1.2, w= 7) & 71.96 & 5.82 & 36/42\\
LODA (IQR=1.2, w=7) & 70.68 & 5.97 & 35/42\\
ABOD (quantile=0.96, w=21) & 52.01 & 3.98 & 26/42\\
\hline
\hline
Multidimensional CMP (k=1, w=7) & 60.33 & 5.58 & 30/42\\
\hline
\end{tabular}
}\vspace{0.25cm}
\caption{Results for Agitation cohort}
\end{table}
\begin{table}
\centering
\resizebox{0.8\textwidth}{!}{%
\begin{tabular}{|p{0.6\linewidth} | p{0.08\linewidth} | p{0.07\linewidth} | p{0.09\linewidth} | }
\hline
Model & Recall\% & Alert rate\% & Patient validity\\
\hline\hline
DOMINANT (alpha=0.9, binned) & 57.97 & 5.53 & 16/23\\
\hline
GNN (Euclidean-Cosine, dim=128, binned) & 60.14 & 4.88 & 16/23 \\
\hline
GCNAE (dropout=0.3, dim=128, layers=4) & 63.77 & 5.60 & 17/23\\
\hline
MLPAE (dropout=0.1, dim=128, layers=1, binned) & 52.90 & 5.47 & 16/23\\
\hline
\textbf{OCGNN (dropout=0.3, dim=128, layers=1, binned)} & \textbf{72.46} & \textbf{5.85} & \textbf{19/23}\\
\hline
\hline
COPOD (IQR=1.2, w=7) & 71.74 & 6.45 & 19/23\\
LODA (IQR=1.2, w=7) & 67.39 & 5.45 & 18/23\\
ABOD (quantile=0.95, w=21) & 40.58 & 4.37 & 12/23\\
\hline
\hline
Multidimensional CMP (k=1, w=7) & 57.25 & 5.55 & 16/23\\
\hline
\end{tabular}
}\vspace{0.25cm}
\caption{Results for Falls cohort}
\end{table}
We choose the top performing model based on recall, for each graph-based anomaly detection technique using a validation set of household movement data collected from the homes of 15 subjects with incidence of urinary tract infection. 
The models are tested on two separate, unseen patient cohorts with incidences of agitation episodes and fall events respectively. 
The models are compared with the best performing CMP, ABOD, COPOD and LODA models included in \cite{bijlani2022unsupervised}. 
Tables 3 and 4 show the results for each cohort. Graph-based models outperform both, CMP and SOTA models, yielding higher average recall, higher patient-level validity and lower alert rate. 
The experimental results confirm that our proposal for graph-based extension is effective in enhancing the performance of CMP-based anomaly detection in noisy real-world sensor data and significantly better than the three SOTA algorithms.  

\subsection{Effect of hyperparameters}\vspace{-0.125cm}
\begin{figure}
\begin{center}
\begin{tabular}{cc}
\bmvaHangBox{\fbox{\includegraphics[width=0.46\linewidth, keepaspectratio]{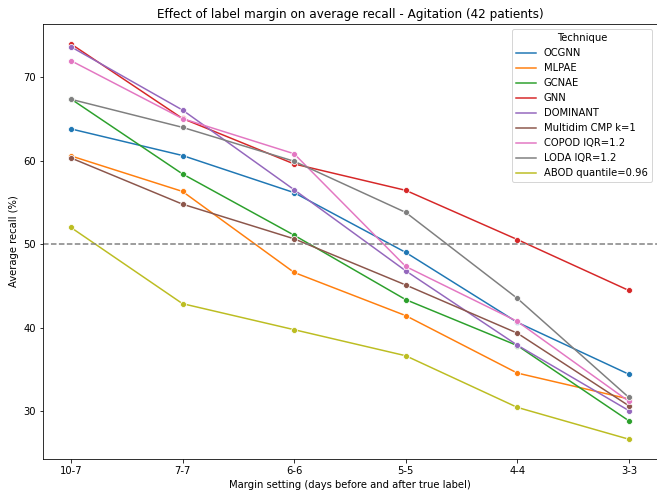}}}&
\bmvaHangBox{\fbox{\includegraphics[width=0.46\linewidth, keepaspectratio]{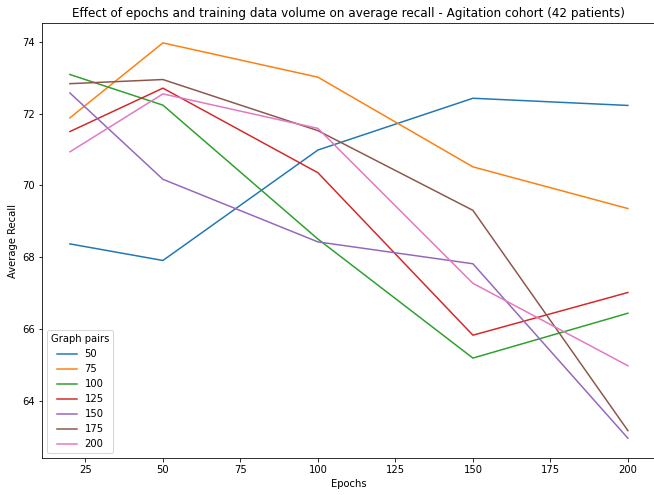}}}\\
(a)&(b)
\end{tabular}
\end{center}
\caption{Effect of label margin on graph models performance}\vspace{-0.25cm}
\label{fig:lab_marg}
\end{figure}

\begin{figure}
\centering
\bmvaHangBox{\fbox{\includegraphics[width=0.7\linewidth,keepaspectratio]{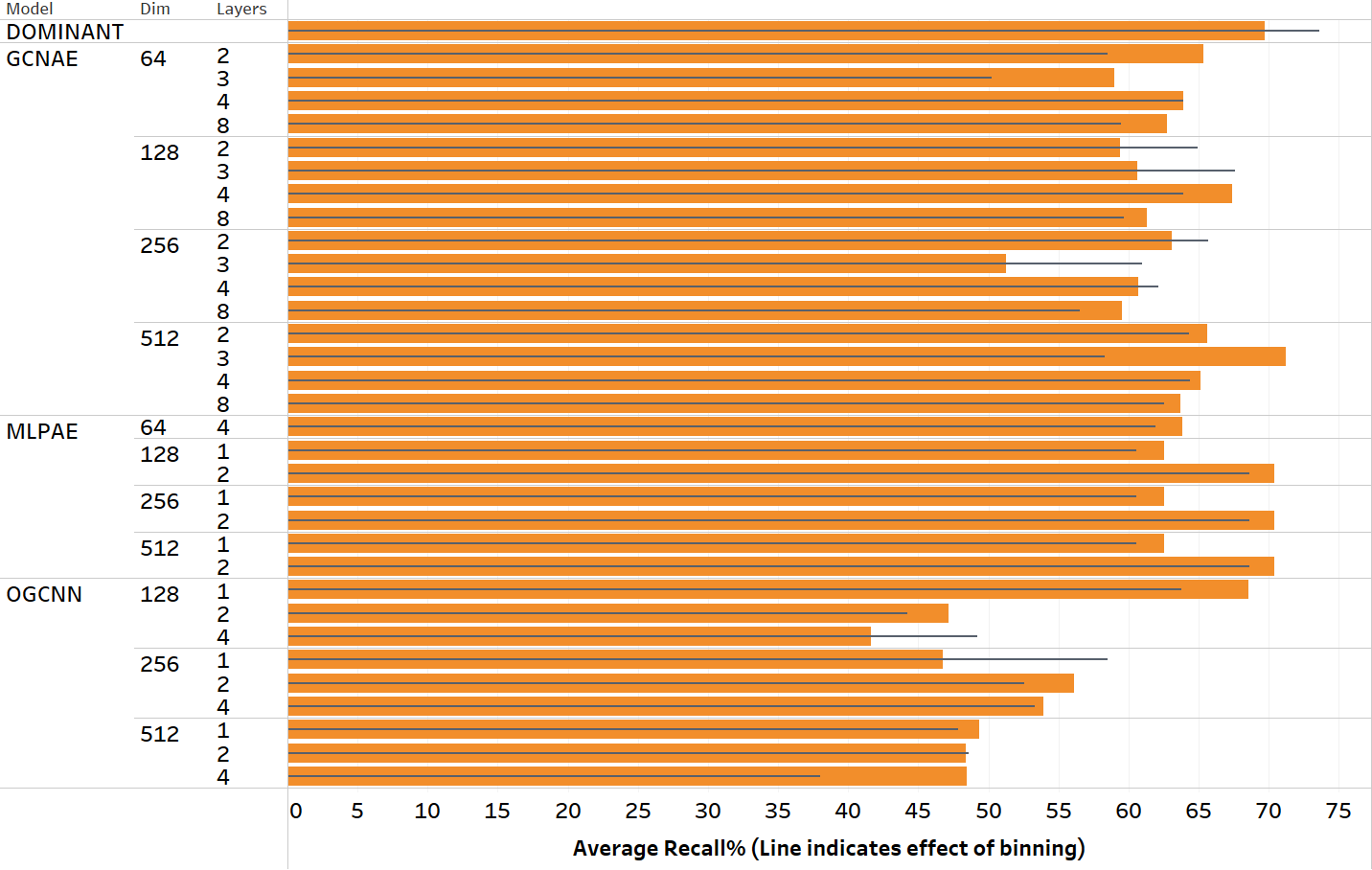}}}\vspace{0.25cm}
\caption{Effect of hyperparameters on graph outlier algorithms - Agitation cohort}
\label{fig:hyper}
\end{figure}
In this section, we use the Agitation cohort data to investigate the stability of the graph algorithms with varying label validation margin settings, effect of coarsening the CMP via binning, and for the self-supervised GNN in particular, number of layers, embedding dimension, and number of training graph pairs for self-supervision. Figure \ref{fig:lab_marg}a shows the effect of varying the margin for label validation from -10, +7 days down to -3, +3 days, on recall. We see here that the self-supervised GNN scores consistently higher than other models across the range of settings. Figure \ref{fig:lab_marg}b shows the effect of the number of training graph pairs and training epochs, on recall performance of the self-supervised GNN. We see that the GNN can achieve optimal recall performance with light touch training using only 75 graph pairs and 50 epochs. More data and continued training hurts performance due to overfitting.  Figure \ref{fig:hyper} shows the effect of coarsening the CMP, number of layers and embedding dimensions on recall performance. Binning the CMP distance matrix before creating context graphs helps improve the performance of DOMINANT and GCNAE to a small extent, but usually has a regressive effect. An embedding size of 128 yields optimal performance across all algorithms. A layer size of 1 works best for MLPAE and OCGNN, while 4 is typically best for GCNAE. No clear pattern emerges when increasing or reducing layer sizes and embedding dimension, however, simpler is often better and the optimal parameters should be chosen based on a dataset with a similar level of noise to the test dataset.
\vspace{-0.125cm}
\subsection{Discussion and Conclusion}\vspace{-0.125cm}
The CMP powered by GNN represents a new generic approach that combines graph-based machine learning with distance matrices for pattern/anomaly detection, of which sensor-based remote health monitoring in healthcare is but one application. This idea can be applied to computer vision, for instance, we could generate the CMP from image representations extracted using a CNN. Graph models add to the power of CMP-based anomaly detection in three ways. First, by design, they allow for complex aggregation of both, temporal as well as cross-feature information when detecting anomalies. This allows graph models to discover patterns and anomalies that may not be detected by using simpler statistical methods. For baseline CMP models, such aggregation must be explicitly defined. Second, graph models allow for great flexibility. Graphs may be constructed in different ways, e.g. nodes can represent time contexts, days or even individual sensors. The anomaly detection algorithm may be conditioned on any suitable measure of anomalousness, e.g. entropy, domain-specific thresholds or cluster-based outlierness. Moreover, graph distance can be defined via custom distance metrics. Third, graph models can create graph embeddings that quantify and encode the anomalousness of graphs. This is useful to understand and analyse specific anomaly types and discover contributing features and patterns. We investigate this aspect in future work. We also note that GNN models outperform autoencoder models, and this is likely due to non-uniformly varying levels of noise in streaming sensor data that affect the latter's performance. Our results show that simple single-layer GNN-based models based on denoising CMPs are capable of outperforming state-of-the-art methods on noisy, real world human activity datasets. The use of CMP as the foundation continues to guarantee high performance and explainability of factors contributing to anomalies. The models achieve high overall sensitivity as well as individual-level sensitivity for over 82\% of our cohort. GNN-based models are also consistently superior when validated at different label margin settings.

In this work, we extended the capabilities of the CMP in unsupervised anomaly detection in noisy sensor-based remote health monitoring, via a novel self-supervised graph anomaly detection pipeline. CMP distance matrices were used to construct temporal context graphs which were then analysed for relative anomalousness based on previous graphs. Validation on two real-world household movement datasets comprising 65 individuals shows that graph-based anomaly detection is more effective than standard CMP and state-of-the-art methods. Our future work includes validation on more sensor-based datasets, more features, constructing graphs in different ways and different guiding functions to improve performance.

\bibliography{egbib}

\begin{thebibliography}{25}
\providecommand{\natexlab}[1]{#1}
\providecommand{\url}[1]{\texttt{#1}}
\expandafter\ifx\csname urlstyle\endcsname\relax
  \providecommand{\doi}[1]{doi: #1}\else
  \providecommand{\doi}{doi: \begingroup \urlstyle{rm}\Url}\fi

\bibitem[UKD()]{UKDRI}
\uppercase{UK} \uppercase{D}ementia \uppercase{R}esearch \uppercase{I}nstitute
  \uppercase{C}are \uppercase{R}esearch and \uppercase{T}echnology.
\newblock \url{https://ukdri.ac.uk/centres/care-research-technology}.
\newblock Accessed: 2022-06-06.

\bibitem[Agata et~al.(2013)Agata, Loeb, and Mitchell]{agata2013challenges}
Erika~D' Agata, Mark~B Loeb, and Susan~L Mitchell.
\newblock Challenges in assessing nursing home residents with advanced dementia
  for suspected urinary tract infections.
\newblock \emph{Journal of the American Geriatrics Society}, 61\penalty0
  (1):\penalty0 62--66, 2013.

\bibitem[Bai et~al.(2019)Bai, Ding, Qiao, Marinovic, Gu, Chen, Sun, and
  Wang]{bai2019unsupervised}
Yunsheng Bai, Hao Ding, Yang Qiao, Agustin Marinovic, Ken Gu, Ting Chen, Yizhou
  Sun, and Wei Wang.
\newblock Unsupervised inductive graph-level representation learning via
  graph-graph proximity.
\newblock \emph{arXiv preprint arXiv:1904.01098}, 2019.

\bibitem[Bijlani et~al.(2022)Bijlani, Nilforooshan, Kouchaki,
  et~al.]{bijlani2022unsupervised}
Nivedita Bijlani, Ramin Nilforooshan, Samaneh Kouchaki, et~al.
\newblock An unsupervised data-driven anomaly detection approach for adverse
  health conditions in people living with dementia: Cohort study.
\newblock \emph{JMIR aging}, 5\penalty0 (3):\penalty0 e38211, 2022.

\bibitem[Cui et~al.(2020)Cui, Zhou, Yang, and Liu]{cui2020adaptive}
Ganqu Cui, Jie Zhou, Cheng Yang, and Zhiyuan Liu.
\newblock Adaptive graph encoder for attributed graph embedding.
\newblock In \emph{Proceedings of the 26th ACM SIGKDD International Conference
  on Knowledge Discovery \& Data Mining}, pages 976--985, 2020.

\bibitem[{De Paepe} et~al.(2020){De Paepe}, {Vanden Hautte}, Steenwinckel, {De
  Turck}, Ongenae, Janssens, and {Van Hoecke}]{DEPAEPE2020103487}
Dieter {De Paepe}, Sander {Vanden Hautte}, Bram Steenwinckel, Filip {De Turck},
  Femke Ongenae, Olivier Janssens, and Sofie {Van Hoecke}.
\newblock A generalized matrix profile framework with support for contextual
  series analysis.
\newblock \emph{Engineering Applications of Artificial Intelligence},
  90:\penalty0 103487, 2020.
\newblock ISSN 0952-1976.
\newblock \doi{https://doi.org/10.1016/j.engappai.2020.103487}.
\newblock URL
  \url{https://www.sciencedirect.com/science/article/pii/S0952197620300087}.

\bibitem[Ding et~al.(2019)Ding, Li, Bhanushali, and Liu]{ding2019deep}
Kaize Ding, Jundong Li, Rohit Bhanushali, and Huan Liu.
\newblock Deep anomaly detection on attributed networks.
\newblock In \emph{Proceedings of the 2019 SIAM International Conference on
  Data Mining}, pages 594--602. SIAM, 2019.

\bibitem[Hu et~al.(2020)Hu, Michel, Russo, Mora, Matrella, Ciampolini, Cocchi,
  Montanari, Nunziata, and Brunschwiler]{hu2020unsupervised}
Rui Hu, Bruno Michel, Dario Russo, Niccol{\`o} Mora, Guido Matrella, Paolo
  Ciampolini, Francesca Cocchi, Enrico Montanari, Stefano Nunziata, and Thomas
  Brunschwiler.
\newblock An unsupervised behavioral modeling and alerting system based on
  passive sensing for elderly care.
\newblock \emph{Future Internet}, 13\penalty0 (1):\penalty0 6, 2020.

\bibitem[Ijaopo(2017)]{ijaopo2017dementia}
EO~Ijaopo.
\newblock Dementia-related agitation: a review of non-pharmacological
  interventions and analysis of risks and benefits of pharmacotherapy.
\newblock \emph{Translational psychiatry}, 7\penalty0 (10):\penalty0
  e1250--e1250, 2017.

\bibitem[Jakkula et~al.(2007)Jakkula, Cook, and Crandall]{jakkula2007temporal}
Vikramaditya Jakkula, Diane~J Cook, and Aaron~S Crandall.
\newblock Temporal pattern discovery for anomaly detection in a smart home.
\newblock 2007.

\bibitem[Kipf and Welling(2016{\natexlab{a}})]{kipf2016semi}
Thomas~N Kipf and Max Welling.
\newblock Semi-supervised classification with graph convolutional networks.
\newblock \emph{arXiv preprint arXiv:1609.02907}, 2016{\natexlab{a}}.

\bibitem[Kipf and Welling(2016{\natexlab{b}})]{kipf2016variational}
Thomas~N Kipf and Max Welling.
\newblock Variational graph auto-encoders.
\newblock \emph{arXiv preprint arXiv:1611.07308}, 2016{\natexlab{b}}.

\bibitem[Kriegel et~al.(2008)Kriegel, Schubert, and Zimek]{kriegel2008angle}
Hans-Peter Kriegel, Matthias Schubert, and Arthur Zimek.
\newblock Angle-based outlier detection in high-dimensional data.
\newblock In \emph{Proceedings of the 14th ACM SIGKDD international conference
  on Knowledge discovery and data mining}, pages 444--452, 2008.

\bibitem[Li et~al.(2020)Li, Zhao, Botta, Ionescu, and Hu]{li2020copod}
Zheng Li, Yue Zhao, Nicola Botta, Cezar Ionescu, and Xiyang Hu.
\newblock Copod: copula-based outlier detection.
\newblock In \emph{2020 IEEE International Conference on Data Mining (ICDM)},
  pages 1118--1123. IEEE, 2020.

\bibitem[Ma et~al.(2021)Ma, Wu, Xue, Yang, Zhou, Sheng, Xiong, and
  Akoglu]{Ma_2021}
Xiaoxiao Ma, Jia Wu, Shan Xue, Jian Yang, Chuan Zhou, Quan~Z. Sheng, Hui Xiong,
  and Leman Akoglu.
\newblock A comprehensive survey on graph anomaly detection with deep learning.
\newblock \emph{{IEEE} Transactions on Knowledge and Data Engineering}, pages
  1--1, 2021.
\newblock \doi{10.1109/tkde.2021.3118815}.
\newblock URL \url{https://doi.org/10.1109%2Ftkde.2021.3118815}.

\bibitem[Mo et~al.(2022)Mo, Peng, Xu, Shi, and Zhu]{mo2022simple}
Yujie Mo, Liang Peng, Jie Xu, Xiaoshuang Shi, and Xiaofeng Zhu.
\newblock Simple unsupervised graph representation learning.
\newblock AAAI, 2022.

\bibitem[Pevn{\`y}(2016)]{pevny2016loda}
Tom{\'a}{\v{s}} Pevn{\`y}.
\newblock Loda: Lightweight on-line detector of anomalies.
\newblock \emph{Machine Learning}, 102\penalty0 (2):\penalty0 275--304, 2016.

\bibitem[Rantz et~al.(2012)Rantz, Skubic, Koopman, Alexander, Phillips,
  Musterman, Back, Aud, Galambos, Guevara, et~al.]{rantz2012automated}
Marilyn~J Rantz, Marjorie Skubic, Richelle~J Koopman, Gregory~L Alexander,
  Lorraine Phillips, Katy Musterman, Jessica Back, Myra~A Aud, Colleen
  Galambos, Rainer~Dane Guevara, et~al.
\newblock Automated technology to speed recognition of signs of illness in
  older adults.
\newblock \emph{Journal of Gerontological Nursing}, 38\penalty0 (4):\penalty0
  18--23, 2012.

\bibitem[Sakurada and Yairi(2014)]{sakurada2014anomaly}
Mayu Sakurada and Takehisa Yairi.
\newblock Anomaly detection using autoencoders with nonlinear dimensionality
  reduction.
\newblock In \emph{Proceedings of the MLSDA 2014 2nd workshop on machine
  learning for sensory data analysis}, pages 4--11, 2014.

\bibitem[Teng et~al.(2018)Teng, Yan, Ertugrul, and Lin]{teng2018deep}
Xian Teng, Muheng Yan, Ali~Mert Ertugrul, and Yu-Ru Lin.
\newblock Deep into hypersphere: Robust and unsupervised anomaly discovery in
  dynamic networks.
\newblock In \emph{Proceedings of the Twenty-Seventh International Joint
  Conference on Artificial Intelligence}, 2018.

\bibitem[Wang et~al.(2021)Wang, Jin, Du, Cui, Tan, and Yang]{wang2021one}
Xuhong Wang, Baihong Jin, Ying Du, Ping Cui, Yingshui Tan, and Yupu Yang.
\newblock One-class graph neural networks for anomaly detection in attributed
  networks.
\newblock \emph{Neural computing and applications}, 33\penalty0 (18):\penalty0
  12073--12085, 2021.

\bibitem[Yang et~al.(2022)Yang, Tian, and Li]{yang2022flow}
Zhengqiang Yang, Junwei Tian, and Ning Li.
\newblock Flow graph anomaly detection based on unsupervised learning.
\newblock \emph{Mobile Information Systems}, 2022, 2022.

\bibitem[Yeh et~al.(2016)Yeh, Zhu, Ulanova, Begum, Ding, Dau, Silva, Mueen, and
  Keogh]{yeh2016matrix}
Chin-Chia~Michael Yeh, Yan Zhu, Liudmila Ulanova, Nurjahan Begum, Yifei Ding,
  Hoang~Anh Dau, Diego~Furtado Silva, Abdullah Mueen, and Eamonn Keogh.
\newblock Matrix profile i: all pairs similarity joins for time series: a
  unifying view that includes motifs, discords and shapelets.
\newblock In \emph{2016 IEEE 16th international conference on data mining
  (ICDM)}, pages 1317--1322. Ieee, 2016.

\bibitem[Zhang et~al.(2018)Zhang, Cui, Neumann, and Chen]{zhang2018end}
Muhan Zhang, Zhicheng Cui, Marion Neumann, and Yixin Chen.
\newblock An end-to-end deep learning architecture for graph classification.
\newblock In \emph{Thirty-second AAAI conference on artificial intelligence},
  2018.

\bibitem[Zheng et~al.(2019)Zheng, Li, Li, Li, and Gao]{zheng2019addgraph}
Li~Zheng, Zhenpeng Li, Jian Li, Zhao Li, and Jun Gao.
\newblock Addgraph: Anomaly detection in dynamic graph using attention-based
  temporal gcn.
\newblock In \emph{IJCAI}, pages 4419--4425, 2019.

\end{thebibliography}
\end{document}